# Complex networks and human language[*]

**Abstract:**

Complex dynamic networks are ubiquitous in nature and human society. In recent years research in large-scale networks has seen a rapid development in various disciplines, inspired by the discovery of two features shared by many real-world networks: small-world and scale-free. This paper introduces how human languages can be studied in this light, and reports some work available in this area. There are two directions of exploration. One is to study networks existing in the language system. Various lexical networks can be built based on different relationships between words, being semantic or syntactic. Recent studies have shown that these lexical networks exhibit small-world and scale-free features. The findings of these global structures of the mental lexicon, which are not detectable in traditional analyses on semantic networks, trigger more interests in the global structures of the mental lexicon and language in general from a network perspective. The other direction of exploration is to study networks of language users (i.e. social networks of people in the linguistic community), and their role in language evolution. Social networks also show small-world and scale-free features, which cannot be captured by random or regular network models. In the past, computational models of language change and language emergence often assume a population to have a random or regular structure, and there has been little discussion how network structures may affect the dynamics. In the second part of the paper, a series of simulation models of diffusion of linguistic innovation are used to illustrate the importance of choosing realistic conditions of population structure for modeling language change.. Four types of social networks are compared, which exhibit two categories of diffusion dynamics. While the questions about which type of networks are more appropriate for modeling still remains, we give some preliminary suggestions for choosing the type of social networks for modeling.

## 1. Introduction

We live in a world full of networks. A network is a system with interconnected components, where the components are called "nodes" and the connections "links". Some systems appear as networks in an obvious way as they have the physical features of network structure, for example, the telephone network, the power and gas transmission system, the Internet, electronic circuits, and so on. More importantly, we can construct networks from systems which do not appear as networks at first sight, as the connections between nodes are not physical links but some abstract relationship, such as food webs, actors' and scientists' collaboration networks, friendship networks, etc. Table 1 gives some examples of networks.

Table 1. Some examples of networks in the real world.

| Networks | Node | Link |
|---|---|---|
| Internet | Computer | cable |
| power network | power station | transmission line |
| food web | Species | predator-pray relation |
| scientific collaboration network | Scientist | co-authorship in papers |
| semantic network | meaning /word | semantic relationship |
| language community | language user | linguistic interaction |

Studies on networks can be traced back to Leonard Euler's mathematical solution of the Königsberg bridge problem in 1735, which started an important branch of mathematics known as graph theory[1].

---

[*] I would like to thank Profs. Ron G-R Chen, John H. Holland and William S-Y. Wang for their support and stimulation for the research reported in this paper. I am also thankful to Dr. James W. Minett and members of the former Language Engineering Laboratory for their helpful discussions. Special thanks are due to Dr. Christophe Coupé and the support of Laboratoire Dynamique du Langage, Institut des Sciences de l'Homme in Lyon, France.

Around 1960, two mathematicians, Paul Erdös and Alfred Rényi, discovered several striking properties of random networks (Erdös & Rényi 1959) and stimulated a bourgeoning interest in the theoretical analysis of networks. Parallel to mathematics, in social sciences, extensive works on social networks have been carried out since the 1930s (Scott 1991/2000). Also, it has been recognized that the network can be the common pattern of life systems at all levels, ranging from food webs studied by ecologists, to the neural systems in the brain studied by neuroscientists which have been applied in computer sciences as artificial neural networks, to the intercellular protein-protein interaction networks and genetic regulatory networks (Newman 2002).

In recent years, network studies have seen another surge of new development[2]. Owing to the availability of powerful computational equipment and techniques, it has become possible to investigate large-scale complex networks in the real world, such as the Internet, World Wide Web (WWW), etc. While early studies mostly carried out analyses of small networks and focused on properties of individual nodes or subnets of the networks, recent studies have been more interested in examining the global statistical properties of large scale networks. Moreover, while early studies mostly assumed either regular or random networks (as illustrated in Figure 1(a) and (b)) as the model for real-world systems, recent empirical studies have found that real-world networks exhibit features which can not be captured alone by either of these two types of networks. Figure 1(c) shows an example of a complex large real-world network — a map of the Internet.

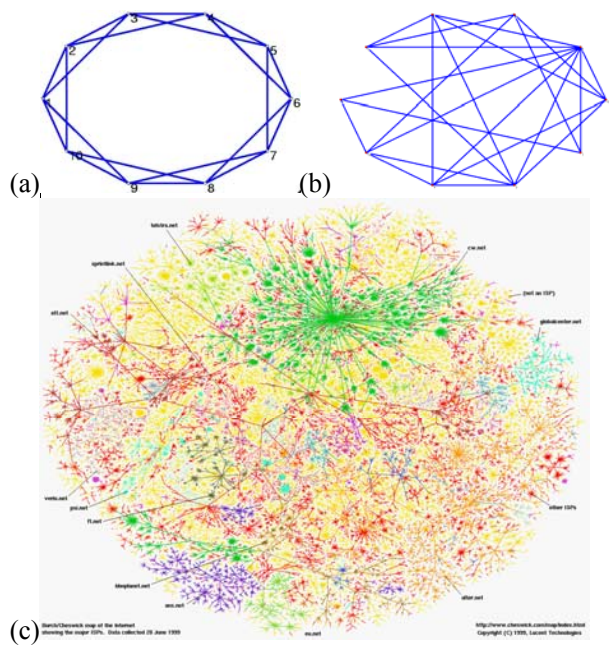

(a) (b)

(c)

Figure 1. Two types of idealized networks: (a) regular network (nodes are regularly connected to the same number of neighbors); (b) random network (nodes are randomly connected)[3]; (c) an example of complex large real-world networks: a map of the Internet as of 28 June, 1999 (from http://www.cheswick.com/map/index.html)

Two important features have been discovered: small-world (Watts & Strogatz 1998) and scale-free (Barabási & Albert 1999). Small-world networks are found, similar to random network, to have a short *characteristic path length* (the terminologies of network measures discussed in the paper are given in the appendix). That is, no matter how large is the network, any two nodes in the network can be connected only through a small number of intermediate nodes. This is what the name "small-world" primarily refers to. At the same time, small-world networks have a high *clustering coefficient*, which is similar to regular networks. In these networks, the nodes are likely to share common neighbors and thus form clusters. Watts & Strogatz (1998) found these two features all present in three networks of different natures, including a collaboration network of film stars, the electrical power grid of the western United States, and the neural network of the nematode worm *C. elegans*. They proposed a simple model (hereafter W&S's model) to build a small-world network by starting from a regular network and rewiring some regular links to long-distance random links.

---

[1] The terms "network", "node" and "link" we adopt in this study are used in computer science. In mathematics, a "network" is called a "graph", and "vertex" and "edge" are the terminologies corresponding to "node" and "link" respectively.

[2] Several popular books and overviews have been published recently for a general audience that provide a good coverage of the background and development of the field (Watts 1999, Barabási 2002, Buchanan 2002, Watts 2003). More technical resources can be found from a number of comprehensive reviews (Dorogovtsev & Mendes 2002, Newman 2003, Barabási & Albert 2003).

[3] The network is visualized using a software package for large-network analysis called Pajek, developed by Vladimir Batagelj and Andrej Mrvar. The package is freely downloadable from http://vlado.fmf.uni-lj.si/pub/networks/pajek/.

Shortly afterwards, Barabási & Albert (1999) reported another striking common feature found in real-world networks, such as the movie actors' collaboration network and power system studied in Watts & Strogatz (1998) and the WWW. These networks, independent of their nature and size, all have a power-law *degree distribution*, i.e. the probability $P(k)$ that a node interacts with $k$ other nodes decays as a power law, following the mathematical relation $P(k) \sim k^{-r}$ (with an exponent $r$ between 2.1 and 4). B&A consider this a "scale-free" feature, which cannot be observed in previous random network models or in W&S's small-world network model, as the degree distribution of both small-world networks and random networks follow a Poisson distribution. An important indication of scale-free networks is the existence of some highly connected nodes (also called "hubs" and "authorities" in directed networks) in the network. B&A proposed a model to generate scale-free networks by incorporating two key features in real world networks, i.e. growth and preferential attachment.

The discovery of the two above new properties in real-world networks has triggered a rapid growth of interests in studying networks in various disciplines. Many real-world systems, which had not been studied as networks, are reformulated and analyzed within a network framework. Many of them are shown to exhibit the small-world and/or scale-free properties. In this light, phenomena which appear as unrelated, such as the spread of AIDS virus, the growth of Yahoo and Google as gigantic hubs in WWW, the Asian economic meltdown, etc., can be explained with the same principles. More importantly, some new insights are obtained for old questions which traditional approaches are not able to achieve. For example, in epidemics studies, when the model takes into account the population structure as a scale-free network, it is shown that the traditional public health approach of random immunization could easily fail. Instead, identifying and immunizing the hubs in the network may provide a more effective way to stop epidemics. Similarly, identifying hubs in cancerous cells and developing ways of taking them out, shows enormous promise in the fight against cancer (Barabási 2002). Meanwhile, more complex measures are applied to large-scale networks, such as network resilience, mixing patterns, hierarchical or community structures, motifs, and so on (Newman 2003). Moreover, the commonality of properties found in the diverse networks also leads researchers to search for explanation for some universal principles governing the formation and evolution of these networks, which has become a topic with increasing importance in network research in general (EuroPhy 2004).

In light of the recent rapid development in network studies in other disciplines, interests in networks related to language are growing as well. Concerning human language, there are at least two types of networks of interest to linguists. The first type is the networks existing in the language system. Network is in fact not a new term in linguistics. The concept of semantic network is said to be traceable all the way back to Aristotle (Anderson & Bower 1973:9). The term "semantic network" dates back at least to Ross Quillian (1968), where he conceives that meanings, or concepts, in language are organized in a network fashion. In semantic networks, words or meanings are the nodes and various semantic relationships are the links. Later development in semantic networks continues to refine the tree-hierarchy representations. The research has been focusing on small scale sub-networks. There had been little concern of the global structure of large-scale semantic networks. In the last few years, researchers have started to construct and analyze lexical networks on a grand scale (including not only semantic but also syntactic relationships), such as in English (e.g. Ferrer & Solé 2001a; Dorogovtsev & Mendes 2001; Motter et al. 2002; Sigman & Cecchi 2002; Steyvers & Tenenbaum 2005) and in Czech, German, and Romanian (Ferrer et al. 2004). It is shown that large-scale lexical networks exhibit interesting features shared with networks in other areas.

The other type of network relevant to linguistics is that of language users, that is, social networks of the members of linguistic communities. Individual language users are taken as the nodes of the network, and the links are the linguistic interactions between individuals. There have been extensive empirical studies on how speakers' language use in an on-going changing situation is correlated with their roles in the social networks (Milroy 1980, Eckert 2000). However, these studies mostly focused on synchronic variations in social networks of small local communities. It is hard to conduct sociolinguistic research on large-scale social networks, and even harder to examine the effect of different social networks on language change on a large time scale. As a complementary approach, computational modeling studies can offer a way to study the long-term effect of language change in speech communities of any sizes and structures. Existing models of language change, however, either do not consider the actual population structure (Niyogi & Berwick 1997), or assume regular or random networks as the population structure (Nettle 1999a). Ke et al. (2004) has shown that different network structures affect the dynamics of language change significantly. To study language change with more systematic and refined network structures, which is mostly an unexplored area, will provide a viable platform to re-examine many theoretical questions in language change and historical linguistics.

In the following sections, we will review some of the existing work in these two directions of investigation of networks and language. Section 2 will introduce a number of studies on large-scale lexical networks, and discuss some questions arising from these studies and possible future directions to explore. Networks of language users will be introduced in Section 3, in the discussion of computational modeling of language change. We will report a series of simulation models for language change in detail, and demonstrate how different social networks affect the dynamics of language change, to highlight the importance and necessity of considering a more realistic structure for the population in the models.

## 2. Network within language

### 2.1    Mental lexicon as a network

One of the features that distinguish human language from other animals' communication system is that human language has a large number of words (Hauser et al. 2002). For a normal high school student, the average size of the receptive vocabulary is over 100,000 (Miller & Gildea 1987). Speakers can navigate the massive mental lexicon in a highly efficient way (the reaction time to judge whether a form is a legitimate word in the language usually takes less than 100 mini-second). How is this possible? It has been recognized that the relational structure, which refers to the fact that words are related to other words, as synonyms, antonyms, hyponyms, or hypernyms, etc, may provide an account for the organization of the mental lexicon. When a word is remembered, a set of properties have to be remembered, such as the phonological form, the grammatical category, the meanings etc. If every word had to be remembered with such a full set of properties, the storage requirement and the processing time would be enormous. However, the relational structure decreases the storage requirement. For example, for those words which belong to a hierarchy of super-ordinate relationship, some of the properties of the super-ordinate words may be inherited by the lower-level words (Quillian 1968). For instance, "animal" has the feature of [+animate] and its hyponyms "dog" and "cat" may inherit this feature without requiring duplicate storage. While early studies on semantic networks mostly focused on refining the local networks of specific semantic domains, the global structures of the network have been largely unexplored. Recently, inspired by the recent new developments in the areas of network theory, scientists have started to examine the global organization of lexical networks in languages.

### 2.2    Recent studies of networks of lexicon

There have been a number of reports about large-scale lexical networks of various nature (Ferrer & Solé 2001a, Sigman & Cecchi 2002, Motter et al. 2002, Ferrer et al. 2004, Steyvers & Tenenbaum 2005, among others). There have been mainly two types of networks, based on the nature of the relationship between nodes considered in the network construction. One is to define the links in terms of semantic relationships, and the other is to define the links in terms of grammatical relationships. Interestingly, despite the different nature of these networks, they exhibit convergent features in their global structures. In particular, the networks all exhibit both small-world and scale-free characteristics. These structural characteristics are considered to reflect the self-organization feature of the lexicon, which may account for the fast retrieval of the mental lexicon. Also, there may exist some universal mechanisms for the formation of these networks, shared by networks in other areas such as biological and technological systems. We now give a brief review of these networks reported so far.

### 2.2.1 Network of synonyms

Motter et al. (2002) construct a network of synonyms using Moby thesaurus dictionary[4]. Words which appear as synonyms in the dictionary have a link between them. It is found that this network shows small-world characteristics. The network is highly clustered (*clustering coefficient* C=0.52), while a random network with the same size and same connectivity only has a C=0.002. The network has a small *characteristic path length* (L=3.16), considering the huge size of the network (30,244 nodes). To connect two randomly selected words, it only needs to go through about 2 other words. Interestingly, similar small L is found in Ferrer & Solé (2001a), which gives L=2.63 for a network with 478,773 nodes, while F&S's model uses a totally different method to construct the network – taking collocation relationships between words as links (to be described in Section 2.2.3). The convergent results from two different methods suggest that the small-world feature is significant, and present in several dimensions of the mental lexicon, both semantic and syntactic.

The degree distribution of the network shows two regimes, an exponential distribution in the part of the lower degrees and a power law distribution in the higher degrees. Based on this observation, they suggest a model different from B&A's model to account for the network formation. A similar two-regime degree distribution is also attested in Ferrer & Solé (2001a) (more details are given in a later section on degree distribution).

Steyvers & Tenenbaum (2005) carry out similar studies on three networks built from different sources: one is based on a free-association database, one uses data in WordNet, and the third one is a synonym network using Roget's Thesaurus. Similar findings have been reported for these networks, including short *characteristic path length*, high clustering *coefficient*, and *degree distribution* following a power law.

### 2.2.2 Network of nouns

Sigman and Cecchi (2002) analyze the structure of nouns in the WordNet database (version 1.6). There are mainly four types of semantic relationships between nouns: I: hyponymy/hypernymy (e.g. "rose" and "flower"); II: antonymy (e.g. "good" and "bad"); III: meronymy/holonymy (e.g. "arm" and "body"); IV: polysemy (e.g. "(department) head" and "head (of a bird)"). In the study, four networks are constructed based on different combinations of different semantic relationships, N1={I+II}, N2={I+III}, N3={I+IV}, and N4={I+III+IV}.

The study carries out three measurements to characterize and compare these four types of network, characteristic path length (L), clustering coefficient (C), and traffic. Figure 2 shows L and C of the four networks, in comparison with corresponding control semi-random networks which are created based on N1 with a number of additional random links to match the total number of links of each networks. It is shown that the characteristic path length of N2 is significantly larger than the characteristic path length of the corresponding semi-random network. In comparison, with the inclusion of polysemy, the networks N3 and N4 exhibit the characteristics of small-world networks, a small characteristic path length similar to a random network and a clustering coefficient significantly larger than random networks. N4 shows that adding meronymy upon the presence of polysemy does not result in any further significant change. Comparing the four networks, we can see that polysemy plays an important role in making the lexical network a small world network.

Furthermore, they investigated the effect of polysemy in the relationship between mean distance and deepness in the hypernymy tree (minimal distance to the root). In the hypernymy tree, the way to go from one meaning to another is climbing up and down the tree. For example, to go from *dog* to *oak*, one would follow the trajectory: dog–canine–carnivore–placental–mammal chordate–animal–life form–plant–woody plant–vascular plant–tree–oak. In the hypernymy tree without polysemy, the mean distance from a node to the rest of the vertices progresses as one goes deeper in the tree. It is found that there is a high correlation between the mean distance and deepness. However, when polysemy is added, this correlation is weakened significantly.

---

[4] The dictionary is available freely from Gutenberg Project Etext (ftp://ibiblio.org/pub/docs/books/gutenberg/etext02/mthes10.zip).

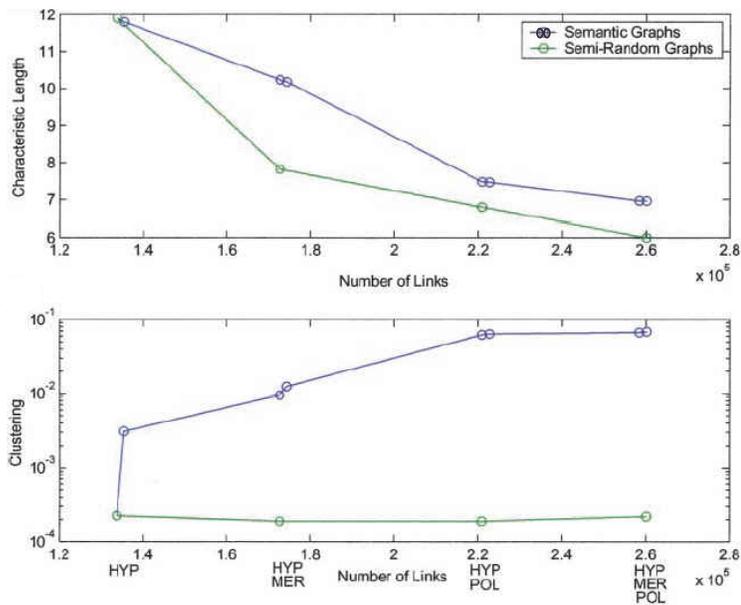

**Figure** 2. Characteristic path length and clustering coefficients of the four constructed networks, compared with semi-random controlled networks (reproduced from Sigman & Cecchi 2002).

They also carried out the traffic analysis, which measures the number of paths passing through each vertex, so as to identify the most active hubs in the network. The study finds that in the network without the presence of polysemy, the most active nodes are those with a large number of hyponyms. This is mainly because all of the corresponding hyponyms have to pass through their hypernymy to navigate to the rest of the network. When polysemy is included, however, the nodes with the most traffic are those words which are the most polysemous and have the most connected neighbors, such as "head", "line" and "point".

### 2.2.3  Collocation network

The above studies mostly deal with the semantic relationships among words. There is another way to construct networks, which is to link words according to their collocation or grammatical relationships. Ricard Solé and his colleagues (Ferrer & Solé 2001a, Ferrer et al. 2004) have carried out a few studies in this direction. In Ferrer & Solé (2001a), two words are linked if they are adjacent neighbors, i.e. they collocate, in a sentence. The British National Corpus, a huge corpus of modern English, is used as the source for network construction. Two types of networks are built: the unrestricted word network (UWN) and the restricted network (RWN). The first takes into account as many collocation links as possible, while the latter considers only those links whose frequency is larger than the chance value (determined by the condition $p_{ij}>p_ip_j$ where $p_i$ and $p_j$ are the probability of occurrence of the two individual words, and $p_{ij}$ is the probability of the collocation. Assuming the words are independent from each other, if a pair of words co-occurs less frequently than expected, the collocation of the pair is considered spurious and is not considered as a meaningful or valid link.

These two networks both exhibit a small world effect: the average distance between any two words is small (L=2~3), irrespective of the huge size of the network. The clustering coefficient (C=0.687 for the UWN and C=0.437 for the RWN) is far larger than what would be expected in corresponding random networks ($C_{randdom}=1.55*10^{-4}$). Furthermore, the degree distribution shows two regimes of power-law. The authors examine further the 5000 most connected words, which they consider as the kernel lexicon[5], and show that the connectivity distribution of these words exhibits a power law with an exponent of — 3.07, which is close to that in the original scale-free network proposed by Barabási & Albert (1999). They speculate that these small-world and scale-free features may be the result of language evolution, to meet the need of optimal navigation in the mental lexicon for speech production.

---

[5] Ferrer & Solé (2001b) discuss the kernel lexicon in greater depth. They argue that the traditional Zipf's law misses the feature of two regimes of frequency distribution in the lexicon, as the corpus used is not big enough (the British National Corpus has about 70 million words, while previous studies only consider texts less than 300,000 words). They show that the distribution can be fitted with two intersecting power-law functions, and estimate the kernel lexicon as the high rank words above the intersection around 5000-6000 words.

The collocation relationship used to build links in Ferrer & Solé (2001a) does not correspond to a syntactic relationship with linguistically precise definition, and fails to capture long-distance syntactic relationships between words. Lately, Ferrer et al. (2004) use some tagged corpora which provide information of head-modifier dependency relationships between words in sentences. Thus they construct syntactic dependency networks, with directed links from modifier to head words. They analyze corpora in three different languages: Czech, German, and Romanian, and show there are strikingly similar characteristics. The networks all exhibit small world patterns (small characteristic path length and high clustering coefficient) and scale-free structure (power-law degree distribution). Also these networks show the presence of a hierarchical organization, as indicated by a high correlation between degree and cluster coefficient. Furthermore, these networks show uniformly a power-law distribution of *betweenness centrality*, and *dissortative mixing* (highly connected words tend not to be interconnected). While these more complex measures need further examination with clearer and more linguistically meaningful interpretation, the highly similar features exhibited in these diverse networks in different languages are very interesting.

For ease of comparison, Table 2 summarizes the results of network analyses from the studies discussed above. All the lexical networks in these studies are all of great size, and they are sparse – the ratios of the number of nodes to the average links are no greater than 0.001. These networks exhibit clearly the two features of small-world characteristics: small distance between words (i.e. small characteristic path length) similar to random networks, and significantly larger clustering coefficients than random networks. These features should play a role in the rapid and efficient navigation in the mental lexicon.

Table 2. A summary of several lexical networks. (N: number of nodes; <k>: average degree; L: characteristics path length; $L_{random}$: characteristic path length of the corresponding random network with same size and average degree; C: clustering coefficient; $C_{random}$: clustering coefficient of the corresponding random network)

| Network | N | <k> | L | $L_{random}$ | C | $C_{random}$ |
|---|---|---|---|---|---|---|
| Motter et al. (2002) (English thesaurus dictionary, synonymy) | 30,244 | 60 | 3.16 | 2.5 | 0.53 | 0.002 |
| Sigman&Cecchi (2002) (WordNet, nouns) HYP | 66,025 | | 11.9 | | 0.002 | $1.2*10^{-4}$ |
| HYP+MER | | | 7.4 | 10 | 0.010 | $1.2*10^{-4}$ |
| HYP+POL | | | 7 | 7.6 | 0.080 | $1.2*10^{-4}$ |
| HYP+MER+POL | | | 6 | 7.2 | 0.081 | $1.2*10^{-4}$ |
| Steyvers & Tenenbaum (2005) (WordNet, all words) | 122,005 | 1.6 | 10.56 | 10.61 | 0.029 | 0.0001 |
| Ferrer & Solé(2001a) (BNC, collocation) | | | | | | |
| Restricted ($P_{ij}>P_iP_j$) | 460,902 | 70 | 2.67 | 3.06 | 0.437 | $1.55*10^{-4}$ |
| Unrestricted | 478,773 | 74 | 2.63 | 3.03 | 0.687 | $1.55*10^{-4}$ |
| Ferrer et al. (2004) Czech | 33336 | 13.4 | 3.5 | 4 | 0.1 | $4*10^{-4}$ |
| German | 6789 | 4.6 | 3.8 | 5.7 | 0.02 | $6*10^{-4}$ |
| Romanian | 5563 | 5.1 | 3.4 | 5.2 | 0.09 | $9.2*10^{-4}$ |

## 2.3 Several questions and ensuing topics

### 2.3.1 Network or not?

The above studies on lexical networks have shown several interesting common features in networks constructed in different ways and across different languages. However, when we probe the actual representation or physical existence of these semantic networks, we find a gap with the connectionist's view on the representation of meanings. In the traditional feedforward and recurrent neural network models, meaning are often represented as the distributed activation patterns of the hidden layer of a neural network, rather than some discrete symbols with physical existence. If meanings do not exist as individual entities, then the structure of the semantic system may not be like a network as the above studies assumed, and the consequent analysis on the networks may not be relevant to the real situation.

To tackle this contradiction, we may consider different levels of representation and analyses. The representation of meanings in the brain may be close to what connectionists have suggested in a distributed manner. But this is the representation at a low level of the neural system. The semantic network may exist as a representation at a higher level. Analyses at different levels of representation can be carried out independently (Jacob 1977), and analyses at a high level of representation may reveal a systems' global features, which could not be obtained when examining the low level representations.

In addition, it will be interesting to develop a method to transform the activation patterns of the neural network into a dynamic network, whose nodes and/or links are changing, and examine the structure of the resultant network. This may provide a framework to investigate the collective global structures of the mental lexicon with the low level representation.

### 2.3.2 The role of polysemy and homophony

The studies reviewed above all show that the lexical network exhibits small-world characteristics, which is taken as an account for the fast and efficient navigation in the semantic network. In particular, Sigman & Cecchi (2002) show that it is polysemy that organizes the semantic network into a small-world structure. Motter et al. (2002), though without explicit claim, seem to have also taken such an assumption, as reflected in the example they use to illustrate the connections in the conceptual network (reproduced in Figure 3(a)). They show there is a shortcut connection between "universe" and "character", through the word "nature". Obviously, "nature" and "character" are polysemes. Meanings which seem distant from each other can be linked through these polysemes. Figure 3(b) gives another example where three distinct meanings, "wrinkle", "cable" and "melody", are linked through the word "line", which is the second most polysemous according to WordNet (the first word is "head"). We hypothesize that the small-world structure found in the synonym network in Motter et al. (2002) is mainly due to the presence of polysemy; when only one meaning of each word is taken to construct the network (to remove the polysemes), the small-world properties may not be present any more. This remains to be tested in further studies.

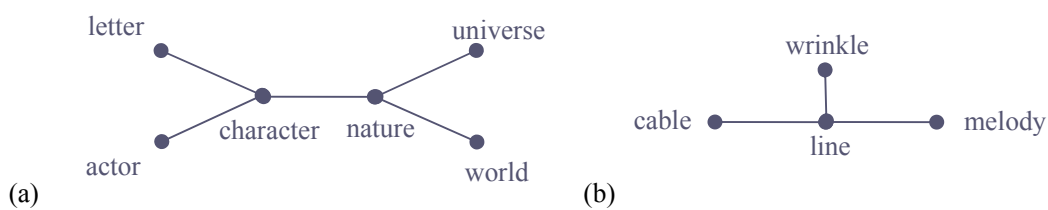

Figure 3. Polysemy provides shortcuts in connecting distant meanings. a) "universe" and ""character" are connected through "nature" (reproduced from Motter et al. 2002); b) "cable", "wrinkle" and "melody" are connected through "line".

Based on the observation that the presence of polysemy reconstructs the semantic network to a small world structure, Sigman & Cecchi (2002) suggest that the advantage provided by polysemy may explain the ubiquity of polysemy across languages. This claim appears to us as confusing the cause and effect. The existence of polysemy is not because fast navigation of the mental lexicon requires its presence to make the semantic network like a small-world network. On the contrary, polysemy is an unavoidable product of human cognition, i.e. from human's metaphoric thinking, imagery, and generalization, which makes semantic extension a universal phenomenon in languages and results in a large number of polysemes (Lakoff 1987). Therefore, the small-world structure of the semantic network is then a consequence of, instead of a cause for, the existence of polysemy.

Furthermore, Sigman & Cecchi (2002) do not make a distinction between polysemy and homophony in their construction of the network, which we consider must necessarily be taken into account. In fact the WordNet database does not provide such a distinction. As polysemy is produced by various mechanisms of semantic extension, different meanings of a polyseme are semantically related, though their relationships may not be obvious when the intermediate stages of semantic extension are not easy to detect after some time, especially when the words are borrowed from other languages. For example, "skate" (as a kind of shoes) and "skate" (as a kind of fish) are both from Old Dutch. In contrast, homophonous words have the same pronunciation but different meanings, and these meanings do not have any semantic relationship, such as "bear" as an animal and "bear" as a verb meaning "tolerate". Most homophones are results of sound change, where two words, originally with different pronunciations, become the same. Therefore, homophony is also an unavoidable phenomenon in language. As homophony and polysemy are distinct in their origin, we speculate that the semantic distance between different meanings of the same form in these two categories would exhibit significant differences. Therefore when these two types of words are differentiated and treated differently in the semantic network, they may change the structure of the semantic network significantly.

### 2.3.3   Degree distribution

In the reviewed studies, most networks exhibit a power law of the degree distribution. However, there are some significant variations. First, the distributions do not all follow a perfect power-law. In both Ferrer & Solé (2001a) and Motter et al. (2002) the degree distributions of the constructed networks exhibit a two-regime feature. Motter et al. (2002) suggest that the high-degree range follows a power-law distribution while the low-degree range shows an exponential distribution, as shown in Figure 4(a). In Ferrer & Solé (2001a) the distribution is fit into two power-law segments (Figure 4(b)).

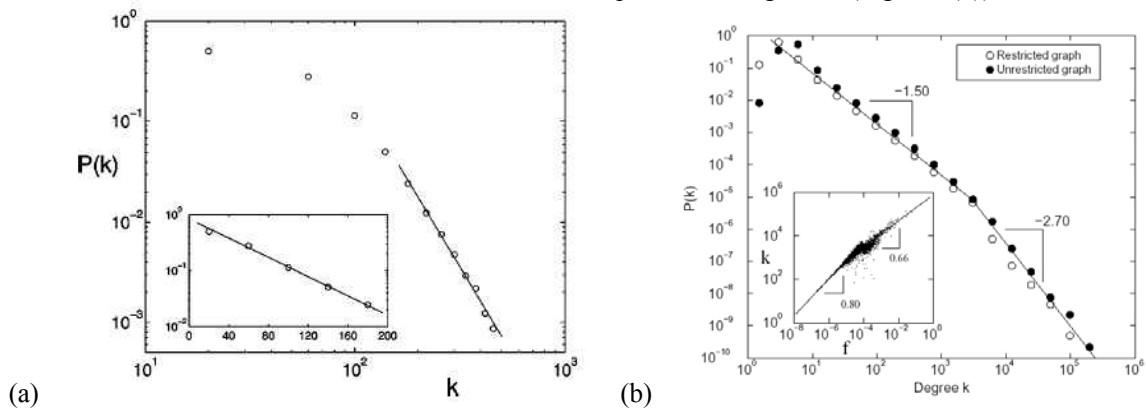

(a) (b)

Figure 4. (a) Degree distribution of words in the synonym network given in Motter et al. (2002). The inset shows the first part of exponential decay in a normal scale; (b) Degree distribution of two networks in the collocation networks reported in Ferrer & Solé (2001a). The inset is the average degree of words as a function of their frequency.

Sigman & Cecchi (2002) show that the three networks constructed respectively with hypernymy-hyponymy, meronymy-holonymy and polysemy relationships all exhibit power-law degree distributions. We carried out similar analysis using a newer version of WordNet (1.7.1) and obtained similar results (Figure 5). However, we notice that in the right part of the graph, the distribution of meanings with high degree become less clean, especially in the network of hypernym/hyponym relationship. There are a small number of senses which are highly connected, such as "herb" ($k=961$), "writer" ($k=896$), "author" ($k=889$) and "bush" ($k=781$), but the distribution of these words does not follow a power-law any more. One account for this irregular part is that these nodes of high degree are constructed with more detailed coverage. For example, the many connections to the sense "herb" include names of different herbs, and "writer" and "author" have many names of the individual writers. So these especially biased nodes make the tail part deviate from a power-law.

Similar explanation can be applied to those degree distributions from the polysemy and meronymy networks. Though these two are less prone to the imbalance, there are some senses which are given a higher degree of elaboration in the definition, compared to other senses. From this observation, we can see the analysis of the global structure of the networks may give a better picture of the whole database of WordNet, and detect some imbalance and bias in the construction (Ferrer et al. (2004) show similar insights for database constructions from their analyses of three syntactic networks). On the other hand, as we are aware of the limitations of the database, we may need to sort out such bias when we make use of the network for further studies or make interpretation of the results.

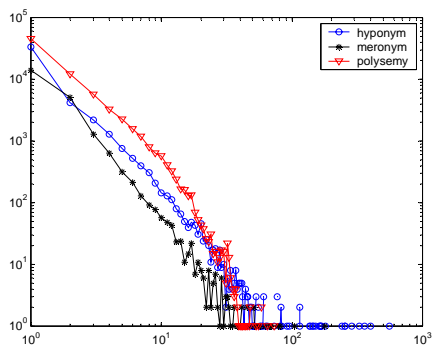

Figure 5. Degree distribution of three types of networks using three different semantic relations provided by Wordnet (version 1.7.1).

### 2.3.4 Collective network vs. individual network

It has to be noted that the above lexical networks were all constructed with the data of a collective lexicon. Either the WordNet, the thesaurus dictionary, or the British National Corpus. In each case the data come from a group of linguists who created the dictionary, or from a large number of speakers from various backgrounds. Meanwhile, we know that different individuals have very different organizations of their mental lexicon. This can be illustrated from the result of a word association experiment, in which subjects come up with different associations for a given word as an immediate response[6].

Table 3 lists the ten most common responses to the word "butterfly" and "hungry", from an early experiment reported in Jenkins (1970). We can see the great diversity in the table. For some subjects, "butterfly" has the strongest association with "moth", a near-synonym of "butterfly"; for some others, the hyponym "insect" is associated more strongly; antonym "full" is recalled first for the word "hungry" in some subjects, whereas some collocation relationships are called up, such as "butterfly"- "pretty" and "man"- "hungry".

Table 3. A list of the top 10 responses for two words "butterfly" and "hungry" from a group of subjects in a word association experiment (Jenkins 1970) (reproduced from Aitchison (1987/1994: 84)).

|   | BUTTERFLY | HUNGRY |
|---|-----------|--------|
| 1 | moth | food |
| 2 | insect | fat |
| 3 | wing(s) | thirsty |
| 4 | bird | full |
| 5 | fly | starved |
| 6 | yellow | stomach |
| 7 | net | tired |
| 8 | pretty | dog |
| 9 | flower(s) | pain |

---

[6] The basic procedure of the experiment is one simple question such as: "give me the first word you think up when I say 'hammer'". Different studies of the word association experiment provided very divergent results, resulting from lack of control. In later experiments with better control on stimuli frequency, reaction time constraint, etc., more convergent findings have been obtained (Aitchison 1987) .

| 10 | bug | man |

Therefore, it would be interesting to compare individuals' lexical networks, and compare the individual networks with the existing collective ones. This may provide some new quantitative measures of the stylistic differences of different speakers. Also, by comparing how much overlap there is between individual speakers' lexical networks, we may get some idea whether there is a similar coverage of the semantic network in individuals. And by comparing how much the individual speakers' lexical networks overlap with the collective semantic network, we may have a better way to examine how well the semantic areas are covered in individuals' limited lexicons. These data may provide some suggestions regarding the question whether the upper limit of the size of individual lexicons (Cheng 1998) is due to the satisfaction of individuals' communication needs, instead of memory limitation.

**2.3.5  Growth of the networks**

Following the analyses of various properties of static networks, the next interesting question is how these networks grow from scratch to the current state. Network growth has become a topic with increasing importance in network research in general (Dorogovtsev & Mendes 2002, EuroPhy 2004). Currently the studies of network growth mostly focus on finding a mechanism which can result in a network with a certain degree distribution. Barabási & Albert (1999) propose preferential attachment (the existing nodes with more links are more likely to be connected with new nodes, i.e. "the rich gets richer") as the growing mechanism for a scale-free network with a power-law degree distribution.

The growth of lexical networks is of particular importance, as individual language users keep increasing their lexicon through their whole life (Sankoff & Lessard 1976), unlike other parts of the language, such as the sound system and grammatical system, which are less likely to change after a certain period, usually around puberty. Furthermore, we are interested in how children develop their lexical networks through language learning from very early on.

In the studies of semantic networks, several proposals of network growth have been given to account for the observed degree distributions. To model the two-part distribution found in the synonym networks, Motter et al. (2002) propose a model incorporating a random attachment in addition to the preferential attachment mechanism. As they explain, "when a new node is added to the network, it has the same probability of attaching to any one of the already existing nodes. But, once it attaches a node *j* it has the tendency to connect preferentially to the nodes that are already connected to *j*". Therefore, the random attachment allows a new node to make a random connection, which provides an equal chance for those nodes with small degrees, resulting in the exponential decay of the low-degree portion in the distribution. Meanwhile, the preferential attachment makes the hubs emerge and results in the power-law in the high-degree portion.

Also starting from the traditional preferential attachment mechanism, Dorogovtsev & Mendes (2001) propose a modified mechanism to account for the two-regime distribution attested in Ferrer & Solé (2001a)'s study. In addition to adding links between the new nodes and the old nodes, they propose that the old nodes themselves may increase their connections gradually. In each step, while the new node is attached to an old node based on preferential attachment, there are a number of new links between old nodes in the growing network. They show that using this method the simulated network exhibits the same two-regime degree distribution found in the empirical network.

Steyvers & Tenenbaum (2005) introduce a growing network model which incorporates two factors in the preferential attachment: new words (concepts) preferentially attach not only to existing highly connected words, but also to words with high utility in terms of word frequency (i.e. words of a higher frequency have a higher utility). The higher the frequency of a word, the higher probability it will get connected. Furthermore, once such a preferential word x is selected for the new word to connect with, the new word is also preferentially connected to the neighbors of x, so as to preserve the local neighbors. The latter mechanism is very similar to what has been proposed in Motter et al. (2002). Also similar mechanisms has been proposed for social network models (Jin et al. 2001), in which new connections are more likely to be made between nodes which have some existing shared neighbors. Figure 6(a) shows the simulation results for the degree with different utilities acquired at different times. They further test the hypothesis using empirical data, including rating of words' acquisition time given by adult and results of children's picture naming, comparing the degree of words with respect to the estimated age of acquisition of the words and frequency of words (shown in Figure 6(b)). It is found that the empirical data confirm well with the prediction shown in Figure 6(a).

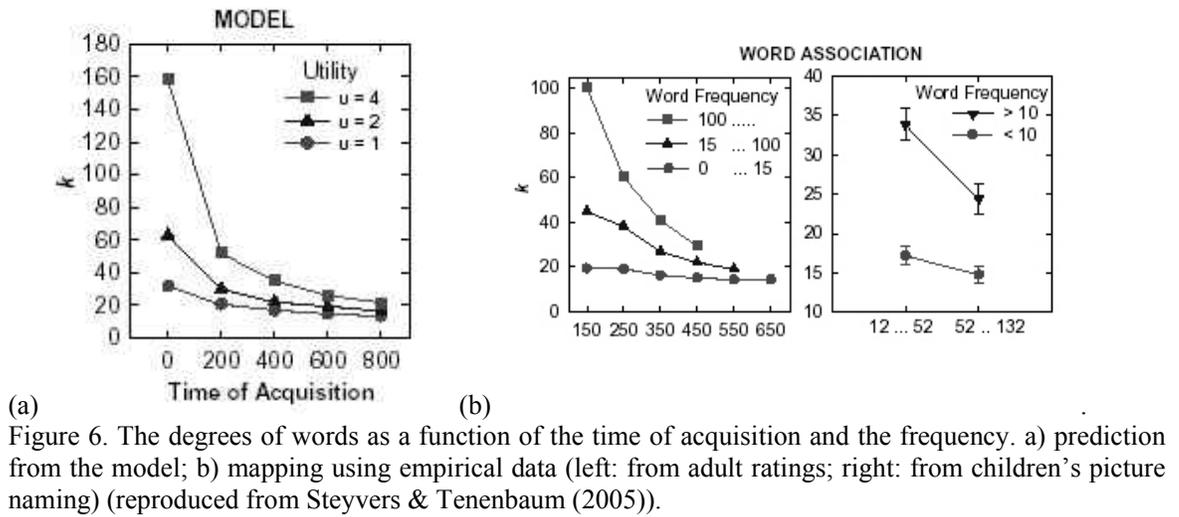

(a)　　　　　　　　　　　　　　(b)

Figure 6. The degrees of words as a function of the time of acquisition and the frequency. a) prediction from the model; b) mapping using empirical data (left: from adult ratings; right: from children's picture naming) (reproduced from Steyvers & Tenenbaum (2005)).

It will be interesting to apply similar analysis as Steyvers & Tenenbaum to more empirical data of children's lexicon development. We may measure the development by comparing the lexical networks of a child at consecutive stages, as well as networks among different children, and networks between children's and their care-takers'. While research has shown that there is a large degree of individual differences in language development, network analyses provide more quantitative indices (Ke & Yao, submitted). For example, when we measure two independent parameters during the growth of the networks, that is, the size and the connectivity, we find that children differ in their developmental paths significantly: children with a small vocabulary (i.e. small network size), often viewed as late talkers, may use the words they have learnt in a more flexible and effective way, which is shown by the large connectivity of their lexical networks. Therefore the traditional simple categorization of early vs. late talkers appears to be insufficient in reflecting these different paths of development. Also, the network analyses provide more informative measures to see how the children's language approaches to the adult model, and how their development interacts with their adult care-takers' language. To illustrate the above observations, Figure 7 shows four children's language development in terms of the two measures on the lexical networks (i.e. size and average degree), and those of their mothers, as reproduced from Ke & Yao (submitted).

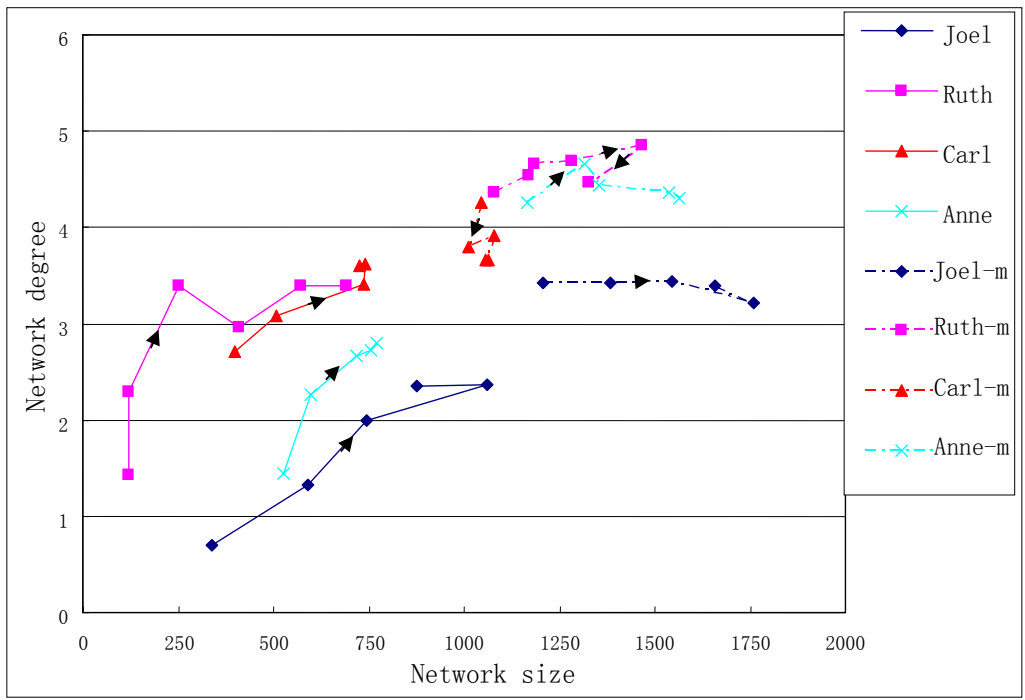

Figure 7. The development of four children's lexical networks in terms of size and average degree, represented by solid lines on the left of the figure, and the corresponding measures of the lexical networks of their mothers, represented by dotted lines on the right of the figure (reproduced from Ke & Yao (submitted)).

It will be an exciting new research direction to analyze language development and carry out cross-linguistic comparison from a network perspective, by making use of the existing large corpora of children's acquisition data, such as CHILDES (MacWhinney 2000). Network analyses can be done independent of specific syntactic frameworks. The global patterns in lexical networks can reveal developmental features which are not easily detected in traditional text analyses, and the syntactical development may be revealed by changes of the global structure of the lexical networks. Network measures may also help to identify individual differences in a more systematic way.

### 2.3.6 More future works

Much interesting future work can be carried out. The construction of lexical networks can be refined in various ways. One interesting finding from word association experiments is that the relationships that show up the most frequently are co-ordination and collocational relationships. This is also observed in lexical decision tasks, tip-of-the-tongue (TOT) phenomena and aphasia (Aitchison 1987/1994). Co-ordination includes not only words which are synonyms and antonyms, but also words which belong to the same hypernym, such as "hungry", "thirsty" and "tired" shown in Table 1, and the color terms "red", "white" and "blue". However, such relationships have not been captured in either WordNet or any thesaurus. It would be worthwhile to examine how the inclusion of such relationships affect the structure of lexical networks.

The networks considered in the current studies are static and the connections are treated as equally weighted and undirected. However, some links may be stronger than others, and some links must be directed, especially in syntactic networks where the directions are very important. Analysis on weighted networks is relatively new in general network theory as well, and recently there have been some available methods which we can make use of (Newman 2004). Moreover, some relationships, such as co-ordination, hypernymy and hyponymy, may be firm connections, while others may be worked out by performing a quick analysis on the spot. For example, "female" is linked with "sow", "princess" and "mare" (Aitchison 1987/1994:95). Therefore, further studies on mental lexicon may consider a dynamic structure, i.e. the links may change from time to time.

Various aphasia may be interpreted from the network perspective. For example, Ferrer & Solé (2001a) show that the most highly connected words in the collocation network are functional words. The deletion of such nodes may change the network navigation dramatically and may account for agrammatism, a kind of aphasia in which function words and bound morphemes are particularly omitted and the speech is non-fluent and labored. When omitted function words are substituted by other words, a syndrome called paragrammatism, the speech will recover fluency in discourse (Caplan 1992). It will be interesting to model this substitution process in the lexical network.

Network studies may generate some new insights and questions for psycholinguistic studies as well. For example, the hubs identified in the various networks may display distinct and recognizable physiological features, as well as a statistical bias for priming in word association and related tasks. Also, lexical access time may be compared and correlated with the semantic distances in the networks.

## 3. Network of language users

Besides semantic networks discussed above, there is another type of network which has been extensively studied in linguistics, that is social networks. Social network has been considered as a determining factor in language change, language contact, language maintenance and shift, etc. (Labov 2001, de Bot & Stoessel 2002). In sociolinguistics, empirical studies in social networks often examine in detail the networks of small communities and focus on the relation between individuals' social network properties and their linguistic performance. A classic study in this area is the work done by Milroy and colleagues (Milroy 1980/1987), in which they examined three stable inner-city communities of Belfast in Britain, and found that the working class communities have "closeknit" social networks in common. The networks are of high density and multiplex[7]. It has been shown that individuals vary in their degrees of integration into the community; some have very few links with people outside their social group, while some have fewer links within the group but more links with the outside. Studies have shown quantitatively that the individual speakers' linguistic behaviors are highly correlated with their degrees of integration into the network: in the situation when linguistic variations are present in the community, the more integrated one individual is into the community, the less variation he has, and the better he conforms to the speech norm of the community.

However, these empirical studies mostly focus on the synchronic linguistic variations in small communities. There have been few studies showing how different social networks affect language change at a larger historical scale. In fact, in the study of language change, social network has created a paradox: while intuitively one would think that the social network should be an important factor in determining language change, very few empirical studies have been able to show the effect of the social network quantitatively over long periods of time (de Bot & Stoessel 2002).

Computer simulations provide us with a convenient platform to study the effect of the social network systematically, as we can manipulate various parameters, such as size, structure of the network, etc. We can simulate the evolution of the network under control conditions, and address questions of language change at different time scales. A micro scale can be adopted to study a given social network in detail to examine its relationship with synchronic linguistic variation. Computational models are particularly useful in addressing questions at a larger time scale, such as how language changes progress through generations in different populations with different social structures, whether the structure of social networks affect the rate of language change, etc. While it is hardly possible to get a clear picture of the structure of a large community at present, and impossible to know in detail what were the social structures in the past, computer models provide us an alternative platform to address these issues, test existing hypotheses, and generate new insights.

---

[7] A network being "multiplex" refers to the fact that two people in a network may have multiple relationships. For example, they could be relatives, neighbors, friends, and/or colleagues.

## 3.1 Language change as a diffusion process

Before going into the discussions about how social networks affect language change, we will consider some earlier models for language change. Language change can be simplified as a diffusion process of some new linguistic elements, i.e. linguistic innovations, in the language community (Shen 1997, Nettle 1999a). The process can be divided into two sub-processes, that is, "innovation" and "diffusion" or "propagation" (Croft 2000)[8]. Here we focus on the second sub-process, the propagation (or diffusion), while the source of innovation is not addressed and its presence is taken as the assumed starting point for the diffusion process.

Traditionally the dynamics of a language change is idealized as an S-curve pattern (Weinreich et al. 1968, Chen 1972, Bailey 1973, Ogura & Wang 1996), i.e. a gradual slow-fast-slow dynamics. Many studies have shown that the empirical data can be fitted with an idealized S-curve, both for sound changes (Shen 1997) and syntactic changes (Kroch 1989, Denison 1993, Ogura 1993, Ogura & Wang 1994). Figure 8 (a) and (b) give two examples reproduced from Shen (1997) and Ogura (1993).

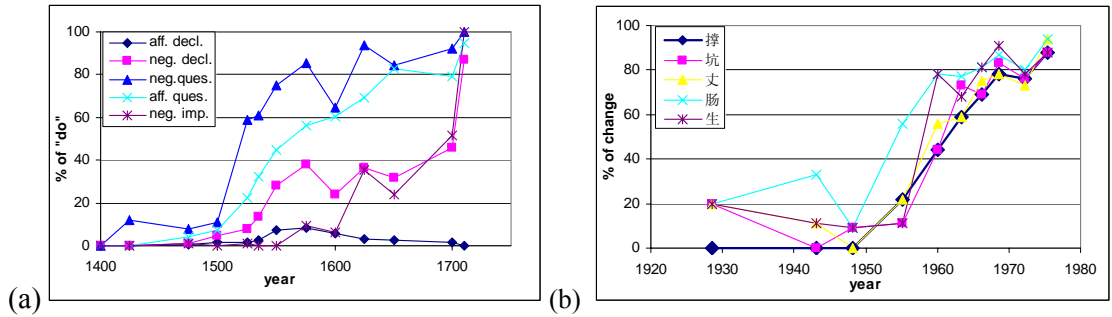

Figure 8. Examples of S-curve from two empirical studies. (a) Development of periphrastic "do" (reproduced from Ogura 1993); (b) diffusion in the apparent time of sound change in Shanghai (produced using data from Shen 1997).

Various mathematical models have been proposed to generate the S-curve diffusion dynamics in language changes (Labov 1994, Shen 1997, Cavalli-Sforza & Feldman 1981). For example, Shen (1997) adopts a model from epidemics, as shown in Equation (1) below, to model a linguistic diffusion process. The rate of increase in the number of people adopting the innovation $C$ (the changed form), is dependent on $c(t)$, the number of people who have already adopted $C$, and $u(t)$, the number of people who are still using $U$ (the original unchanged form), and the *effective contact rate*, α, which is dependent on both the individual's propensity to change and the learnability of the innovation.

$$(1) \quad \frac{dc}{dt} = \alpha\, c(t)\, u(t)$$

While this analytical model is able to produce the logistic diffusion dynamics observed in empirical data, however, some assumptions taken by the model are unrealistic, for example, infinite population size, static population without generation progression, homogeneous individuals in terms of contact rate and learning capacity, and so on. When such a deterministic model is simulated as a stochastic one, the conclusion generated from the deterministic model become invalid as we will show below. Similar criticisms of deterministic models can be found in Briscoe (2000) in his review of the models in Niyogi & Berwick (1997).

First of all, in the deterministic model given by Equation (1), the innovation always diffuses. However, in reality, while linguistic innovations arise frequently in the speech community, only a small number of them are successful in leading to a change. There is a "threshold problem" for successful changes (Nettle

---
[8] There have been similar proposals of dividing the process of language change into two sub-processes, for example, Weinreich et al. (1968)'s "actuation" and "transmission", and Chen & Wang (1975)'s "actuation" and "implementation", though there are minor differences in the definitions of these terms.

1999a,c): why a new linguistic variant, initially rare, can win over the previous linguistic norm? In biological and cultural evolution, when a new mutant trait arises in an individual, it has a good chance to be passed on to that individual's offspring, as long as the mutant is not severely deleterious or actually lethal. Some genetic mutations can successfully diffuse into the whole population without natural selection (Kimura 1983). But linguistic transmission is different from genetic transmission. Instead of inheriting genes from *one or two parents*, a language learner samples at least a proportion of the language community, which may include *a fairly large number of people* in the generations above him as well as in his peer group. The mutant who arises in the last generation will be the minority and unlikely be learned by the next generation. "All plausible learning algorithms lead, other things being equal, to the adoption of the most common variant in the sample for a given item, which will never be the new mutant" (Nettle 1999c: 98). Therefore, new mutants cannot become the fixed norm in a language community unless "they can pass a threshold of frequency which in the early stages they never have" (ibid).

Nettle (1999a, c) suggests that there are two possibilities for the innovation to overcome the threshold. One is functional selection, i.e., there is a *functional bias* toward the innovation over the original norm. Studies on language universals and language evolution have proposed various functional accounts, such as perceptual salience, production economy, markedness, iconicity, etc. (Croft 1990, Kirby 1999). The other possibility to cross the threshold for change is "social selection", in which the innovation originates from some influential speakers who have higher influence, or "social impact", than others and learners may favor learning from them.

Nettle's model is an adapted version of the Social Impact Theory which simulates attitude change in social groups (Nowak et al. 1990). The population is structured with age and social status. The language learner chooses one of the competing linguistic variants by evaluating their impact after sampling the speech of individuals in the community. Individuals within shorter social distance or with higher social status have a higher impact on the learner.

Nettle's simulation models demonstrate that in a community homogeneous in social status, the functional bias needs to be unrealistically high for the innovation to spread successfully; but with social selection, for a population with large differences in social status, an innovation with a very small functional advantage has a high chance to spread. Concluding from these results from the simulation, Nettle suggests that functional biases may affect the direction of language change, but cannot provide a sufficient condition for change to happen. He remarks that "without the potential for change provided by differences in social influence, functionally favored variants might never overcome the threshold required to displace prior norms" (Nettle 1999c: 116).

The conclusion given in Nettle's model that language change requires the existence of super-influential agents to spread the innovation seems too constrained. The hypothesis may find difficulty in explaining "changes from below", where the change diffuses, not from the highest social class, but from the upper working class or lower middle class (Labov 2001: 31).

In the following, we will go through a number of simulation models, starting from a highly idealistic implementation, and adding more and more realistic conditions into the model, to demonstrate how the conditions for language change may be modified when we refine the assumptions in the models.

## 3.2 The models

### 3.2.1 The idealized situation

We will start with a very simple analytic model defined by Equation (1), which is a logistic function and produces an S-curve diffusion dynamic, as shown in Figure 9(a). We transform this highly idealized analytical model into a stochastic multi-agent simulation model, with the following assumptions: the population is constant, static, and finite; the agents are in a fully connected network so that any two agents can interact; the change is unidirectional (only $U \to C$, no $C \to U$); all agents[9] have the same probability α (set as 0.5 here) of adopting the innovation $C$. In the following simulations, the population size is set to N=400 unless specified otherwise. The number is chosen for comparison with Nettle (1999a)'s model.

In the beginning, one random agent creates the innovation $C$ while all others have $U$. Agents start to interact; at each time instant, two agents are selected. If one agent has the innovation, the other agent who still uses $U$ will have a chance probability (α=0.5) to change to $C$. After a number of interactions, the innovation $C$ diffuses to the whole population, exhibiting an S-curve dynamics. As there are random parameters, such as which two agents are selected to interact at a given time, whether the agent changes or not during one interaction, and so on, the actual diffusion processes in different runs show small variations; but they all show S-curve patterns. Figure 9(b) shows the results of 10 runs. Each curve in a graph traces one diffusion process. The average time for the diffusion to complete is about 4,000 steps.

### 3.2.2 Removing full connectivity

In reality, networks are very rarely fully connected, even when very small. In the study of social networks, it is believed that there is an upper limit to the number of relations that each agent can sustain, constrained by time and cost of making and maintaining relations. The maximum value for density[10] of connections which is likely to be found in actual social network is suggested as 0.5 (Scott 1991/2000). We examine how the diffusion dynamics change if the connections in the network become sparser, assuming the network is a regular network (the nodes are arranged in a ring and each node connects to a fixed number of close neighbors, as shown in Figure 9(c)).

As expected, the diffusion slows down and the slope of the S-curve decreases. Figure 9(d) shows the situation when the density drops to 50% in the regular networks, i.e. each node only connects to half of the population. Within the given number of 10,000 steps, only 3 runs out of 10 see complete diffusion, and on average the innovation only reaches about 85% of the population. Despite the slow dynamics, the diffusion still proceeds in an S-curve shape.

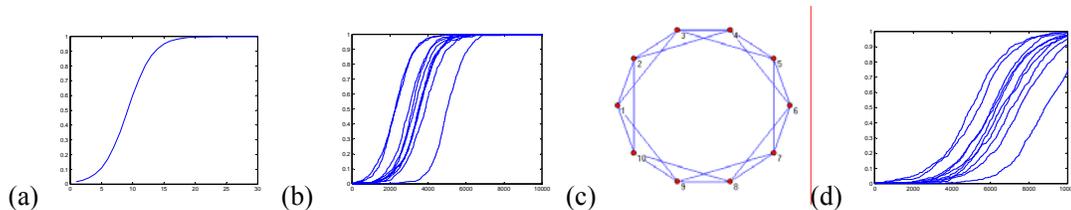

(a)     (b)     (c)     (d)

Figure 9. (a) Idealized diffusion dynamic from an idealized analytic model; (b) Diffusion dynamics in a fully connected network (connection density D=1.0); (c) An example of a regular network; (d) Diffusion dynamics in a regular network (D=0.5). The population size N=400, the change rate α=0.5. One graph includes simulation results from 10 runs. The x-axis is the number of steps, and the y-axis is the proportion of agents using the changed form.

### 3.3 Adding age differences

In the idealized model, the agents are assumed to be homogeneous in their learning capacity, and are immortal. These are unrealistic assumptions. In reality, adults and children may have different responses to linguistic innovations. It is often assumed that adults seldom change their linguistic behaviors after adolescence. The "apparent time" approach (Labov 1994) to reconstructing the historical profile of on-going changes in sociolinguistic studies is based on this assumption. In the following models, we will take into account age differences, and make the population change continuously, similar to Nettle (1999a).

---

[9] In the modeling terminology, we call the individuals in the population "agents". And we may call them "nodes" when we refer to them in the network terminology. In this study, "individual", "agent" and "node" are used interchangeably.

[10] Network "density" is often calculated as the percentage of actual links among the maximum possible links in the network.

In Nettle's model, each agent in the population passes through five life stages. Agents at stage 1 are infants and children, who only learn from others, and do not influence others; agents at stage 2 are adolescents, who both learn and affect others; agents at stages 3-5 are adults, who use what they have learned before adulthood to teach the infants and children, and themselves do not change (in our model, we will allow the adults have a probability to change ($α_{adult}$), in contrast with the children's probability of change ($α_{child}$)). The population is initialized with an equal proportion of agents in these five stages. Thus the ratio between children and adults is 2:3[11]. At the end of each life stage, each agent advances to the next stage, and agents at stage 5 are replaced with new agents at stage 1. The population is structured in a two-dimension grid (it is in fact a spherical shape, with the boundaries connected to each other, to eliminate the edge effect), as shown in Figure 10. This structure represents a social structure with both familial and horizontal social ties incorporated. Agents are close to their parents and siblings in the rows, and close to their peers along the columns of the grid.

```
54321123455432112345
54321123455432112345
54321123455432112345
```
Figure 10. The social structure in a grid representation in Nettle (1999a)'s model.

Due to the incorporation of the age progression and generation replacement, the time step in the simulation becomes an important factor and deserves more careful consideration. In the earlier model, only one pair of agents interact in each time step. This becomes unrealistic if we interpret the time step in the model in terms of age stage. If we divide the life span into five age stages, one age stage is approximately 15 years. Within one stage, each agent should have a large number of interactions with other agents they have connections with. In the following, for the sake of simplicity, we will set a fixed number of interactions for each agent.

Figure 11 shows the diffusion dynamics for different connection densities (D), adults' probabilities of change ($α_{adult}$) and children's probabilities of change ($α_{child}$). The connection density D not only affects the rate of diffusion, as shown in Figure 9, but also affects the shape of the diffusion curve. When D is large (D=0.5 in Figure 11(a) and (b)), the diffusion is highly rapid in a nonlinear way. When D becomes smaller, the curve becomes more linear (compared Figure 11(b) and (c)).

The children's probability of learning the innovation ($α_{child}$) affects the diffusion pattern significantly (compare Figure 11(a) and (b); as well as (d) and (e)). When $α_{child}$ is too small (0.05), the innovation does not always diffuse; sometimes it fails. As shown in Figure 10(e), in 6 out of 10 runs the innovation dies out (they show no trace in the graph), and in the 4 runs where the innovation spreads, it does not reach the whole community — there is a relatively constant proportion of the population (10% as in Figure 11(e)) which keeps the original norm. This incomplete diffusion is interesting and unexpected. A similar phenomenon is found in simulation studies of change of attitudes as well, where incomplete polarization of opinions reaches a stable equilibrium (Nowak et al. 1990). However, we consider this is an artifact of the children's low learning probability and the regular network structure under the current implementation of the learning process. We will see that this incomplete diffusion does not exist in another implementation of the learning process in the next section.

Figure 11(e) also illustrates the non-deterministic nature of language change: starting from the same network and agents with the same probability to change, in some runs the diffusion is successful while in others it is not. The randomness is due to the different origin of the innovation (whether agents at stage 3, 4 or 5 – children are assumed not to innovate here), and the fact that agents change in a probabilistic manner. The small differences in the early stage of the diffusion may accumulate and amplify, and finally result in qualitative differences in the long run – some innovations reach the whole population while some disappear after just a few steps.

---

11 This ratio may not conform to the demography in a modern society. For example, in Hong Kong, the ratio between children (<14 year old) and adults (>15 years old) is about 1:4 (from Hong Kong 1996 census: http://www.info.gov.hk/censtatd/chinese/hkstat/fas/pop/by_age_sex_index.html). However, as the aim of the model is to illustrate the importance of considering non-homogeneous and dynamic populations, the actual ratio is not crucial, and will not affect the quantitative conclusions drawn from this model. Furthermore, this ratio with a smaller difference between adults and children may be appropriate if we consider the pre-modern human society or hunter-gatherer society (Hassen 1981).

Surprisingly, $\alpha_{adult}$ does not affect the success of diffusion, as long as $\alpha_{child}$ is high enough. A smaller $\alpha_{adult}$ only makes the diffusion slower (Figure 11(c) and (d)). As a comparison, when $\alpha_{child}$ is small, even though $\alpha_{adult}$ is high, the innovation can only spread to half of the population and be stabilized at that level (Figure 11(f)) (similar to Figure 11(e), the stabilization in the middle is due to the regular network structure. We will see different outcome in other types of networks later). This suggests that if an innovation is more learnable than the original norm for children, then the innovation is very likely to spread to the whole population and become a new norm, no matter whether adults change or not. Therefore, those language changes, such as the loss of final consonants, which are prone to happen in language acquisition, are more frequently attested. On the contrary, forms which are difficult for children to learn are unlikely to lead to a complete change. The latter part of the conclusions, however, may find difficulties in explaining changes which do seem to be more difficult than earlier norms; models which adopt different social networks discussed in later sections will modify this conclusion.

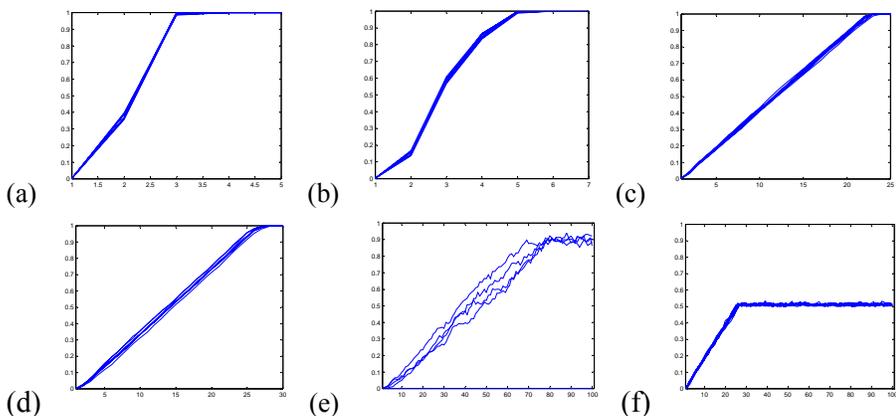

Figure 11. Diffusion in age-structured regular networks. (a) connection density D=0.5, adults' probability of change $\alpha_{adult}$=0.5, children's probability of change $\alpha_{child}$=1.0; (b) D=0.5, $\alpha_{adult}$=0.5, $\alpha_{child}$=0.5; (c) D=0.05, $\alpha_{adult}$=0.5, $\alpha_{child}$=0.5; (d) D=0.05, $\alpha_{adult}$=0.001, $\alpha_{child}$=0.5. (e) D=0.05, $\alpha_{adult}$=0.001, $\alpha_{child}$=0.05. (f) D=0.05, $\alpha_{adult}$=0.5, $\alpha_{child}$=0.001.

### 3.4 Different types of networks

The above model adopts a regular structure for the social network. However, real social networks hardly look like regular networks. Though people are more likely to connect to their family members, local neighbors, colleagues, and so on, they also have long-distance connections such as relatives, friends, or business partners living in other places, or in very different socioeconomic classes of the society. Moreover, people do not have the exact same number of links; some may have more than others. At the same time, real social networks are not like random networks either, as most people would have more local regular connections than distant connections. Recently, it has been found that large-scale social networks exhibit the small-world characteristics (i.e. small characteristic path length and high clustering coefficient), such as in a collaboration network of film actors (Watts & Strogatz 1998), and scale-free characteristics (i.e. a power-law degree distribution), such as in a romantic relationship network (Liljeros et al. 2001). There have been various ways to construct networks to exhibit these two features (Dorogovtsev & Mendes 2002, Newman 2003; Barabási & Albert 2003). There are some models proposed specifically for social networks in order to capture another important feature in social networks: the presence of sub-communities, i.e. clusters (Jin et al. 2001). However, as a preliminary attempt, in this section, we will only study the two classic ones which were proposed at the time of the discovery of small-world and scale-free characteristics.

The small-world network is built based on the model proposed in Watts & Strogatz (1998), as shown in Figure 12: starting from a regular network, rewire a number of regular links randomly according to a probability *p*. The rewiring probability *p* determines statistically how many numbers of regular connections are changed to shortcuts. Watts & Strogatz (1998) show that the transition from a regular network to a small-world network is almost undetectable: when there is only a very small number of random shortcuts, *p* as small as 0.0001, the network readily exhibits the small-world characteristics. For a range of *p*, the network exhibits both the characteristic of a regular network (a high clustering coefficient *C*) and of a random network (characteristic path length *L* much smaller than that of a regular network).

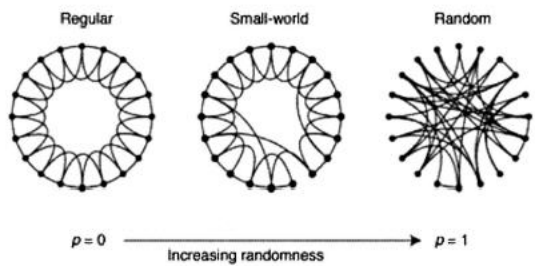

Figure 12. The rewiring method to construct a small-world network, and the interpolation between a regular network and a random network, by increasing the random rewiring probability *p* (reproduced from & Strogatz (1998)).

The scale-free network is built following the growth model proposed in Barabási & Albert (1999). The network starts with a small number of isolated nodes, say $m_0$; at every time step, say the $t^{th}$ time step, a new node is added into the network; each new node gets *m* ($m<m_0+t$) connections, linking to *m* different nodes in the network based on a preferential attachment mechanism: the more connections a node has, the higher the chance for it to be chosen to connect with the new node. After *t* time steps, the procedure results in a network with $N = t + m_0$ nodes, and m*t connections. The final network exhibits the scale-invariant characteristics in its degree distribution, as shown in Section 2.3.3. Figure 13 shows a small scale-free network generated from the algorithm. An obvious feature of the scale-free network is that there exist "hubs" — nodes which are extremely highly connected.

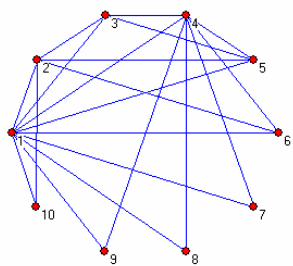

Figure 13. A scale-free network (N=10, average degree <k>=4).

To demonstrate how these different types of network structures affect the diffusion dynamics, we simulate the diffusion in four types of networks, each for 10 runs, with the same set of conditions: network size N=400, average degree <k> = 20 (i.e. connection density D=0.05), and $α_{adult}$=0.001, $α_{child}$=0.5. The networks are always constructed in a way to ensure that there is no isolated node or cluster. The results of the simulation are shown in Figure 14.

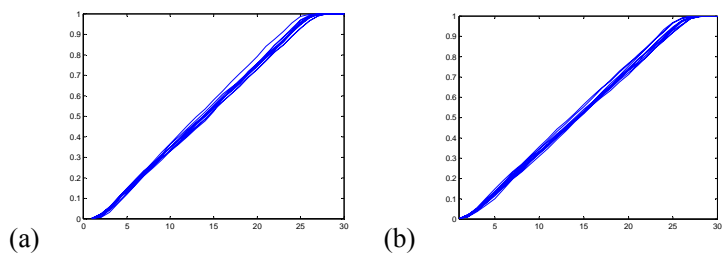

(a)        (b)

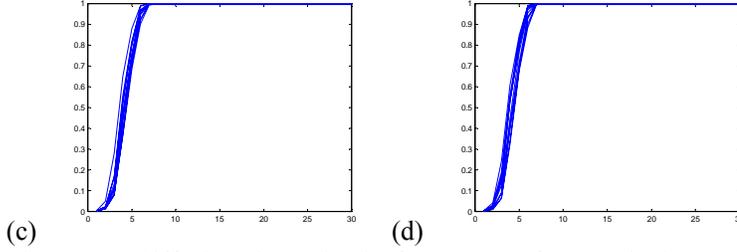

(c)                                        (d)

Figure 14. Diffusion dynamics in four types of networks in 10 runs (N=400, average degree $<k>=20$, $\alpha_{adult}=0.001$, $\alpha_{child}=0.5$). (a) regular network; (b) small-world network (*p=0.01*); (c) random network; (d) scale-free network.

From the above four figures, we can see that the diffusions exhibit two types of dynamics: the regular network and small-world network show slow diffusion in a linear manner, while scale-free and random networks show sharp S-curve rapid diffusion. The differentiation among the four types of networks was not expected. Though the small-world network in the current implementation (the random rewiring rate is low: *p=0.01*) is structurally very close to regular network, we would expect that the small-number of random links could increase the diffusion rate to some degree. The reason for the similarity of results between regular network and small-world network is mainly due to the low connection density. If the connection density D increases, the small-world network will show a faster diffusion rate than the corresponding regular network.

The diffusion rates in the random and scale-free network are both very high (only within about seven generations), and it is hard to see the differences between them. One plausible reason for the fast diffusion rates in these two types of networks is that there exist some nodes, i.e. the hubs, which have a large number of links. Though the existence of such hubs in scale-free networks is much more salient (the degree of hubs is much larger in a scale-free network than in a random network), the similarity of diffusion pattern in these two types of networks suggests that only a small number of nodes with relatively high degrees will suffice to lead to a fast diffusion.

The dichotomy of diffusion dynamics leads to the following two observations concerning how to choose an appropriate social network model for simulation study of language change: 1) regular and random networks can serve as good approximants for more realistic structures of the social networks, under the current implementation of the diffusion of linguistic innovations (the similarity may not be valid when more complex situations are taken into account); however, 2) the question "what is a better model of realistic social networks" still persists. While empirical analyses have shown that some social networks exhibit small-world properties, and some others exhibit scale-free features, so far there has been no well-developed model which can incorporate both features. The model also needs to incorporate the co-existence of local and distant links, and the age structure.

### 3.5 Different functional biases toward the innovation

In the above models, the diffusion of the innovation is modeled from an "interaction" perspective, that is, in each time step, the agents all go through a given number of interactions with their connected neighbors, and each interaction with an innovator gives the agent a chance to learn the innovation. This implementation faces a difficulty in determining an appropriate value for the number of interactions that an agent may have in one time step. We will shift from this "interaction" perspective to a learning perspective.

The learning process is simplified as follows: at each time step, each learner takes input from all his connected "teachers" (any agent older than stage 1). If all teachers use the same form, then there is no doubt that the learner will only learn one form. If both forms, i.e., *U* and *C*, are present in the learning environment, the leaner compares the cumulative impacts of the two variants and chooses the one which has the higher value. This is similar to the implementation in Nettle's model. However, Nettle's model simulates a weighted fully connected regular network which we consider unrealistic: the learner is assumed to be able to have contact with the *whole* population, and teachers' impacts are weighted according to a quadratic or even higher power function with respect to the distance.

Furthermore, in the above models, how the children acquire the innovation is dependent on a probability of change $α_{child}$. In fact, $α_{child}$ is not a very appropriate term, considering the fact that a child's learning of the innovation is different from an adult's: whereas the adult has to give up a form in use and adopt a new form, the child starts from scratch and there is nothing for him to change at the beginning. An adult's change is mostly determined by his sensitivity to the innovation and willingness to change; in contrast, which form a child acquires is determined by the form's learnability in comparison with its competitor, and also the frequency of the two forms.

Therefore, in the following, we will discard the parameter $α_{child}$ (children's probability of change), and instead use another parameter called "functional bias" (β), which is the ratio of the functional values of the two forms: $β=f(C)/f(U)$ ($f(U)$ and $f(C)$ are the functional values[12] for the two competing forms $U$ and $C$ respectively). The functional value for a linguistic item can be considered as a composite abstract index of the form in terms of production ease, perception salience, simplicity and generality of the activated grammatical rules, and so on. Those forms which are shorter and more salient to perceive have a higher functional value and they will be preferred in language acquisition[13].

It is obvious that when the functional bias is smaller than 1 (β<1), i.e. the innovation $C$ has a smaller functional value than $U$, there is no chance for the innovation to spread when it starts in a rare minority. Then, how large should the functional bias β be in order to "beat" the original norm and become a new norm? In the following we will examine the effect of β on diffusion success and diffusion rate in different types of networks. To make the diffusion easier, we set the number of innovators to be 10, instead of 1 in earlier simulations. The initial conditions are as follows: population size N=400, average degree <k>=20, adults rarely change ($α_{adult}$ =0.001). Figure 15 shows the diffusion dynamics in the four types of network when the functional bias β=10. Regular networks and small-world networks show similar gradual diffusion, while random network and scale-free network still exhibit the rapid diffusion. The diffusion always completes, reaching the whole population, as long as the functional bias is strong enough. The incomplete diffusion, which happens in the model of regular network discussed in Section 3.3, does not exist in the current model when a more realistic learning process is implemented.

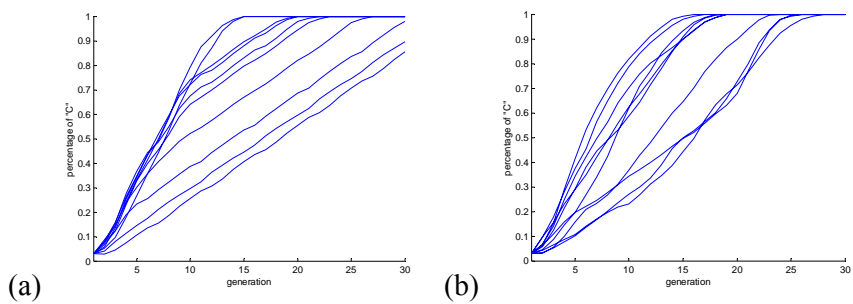

(a)          (b)

---

[12] Functional values $f(C)$ and $f(U)$ correspond to the "acquisition bias" used in Nettle (1999a).

[13] Here is one example of innovation having a higher functional value. It is the serial names for SARS during its outbreak in 2003 in the Chinese mass media. At the beginning the name was a complete translation of the technical term in English: "紧急严重呼吸道疾病症侯群" (Severe Acute Respiratory Syndrome), but this name was quickly overcome by a more common name "非典型肺炎" (Atypical Pneumonia); this term was then further shortened as 非典肺 and later 非典 in mainland China, or 沙士 (from the English abbreviation SARS) in other Chinese communities such as Hong Kong. Obviously, the shorter names are assumed to have a higher functional value over the proceeding longer ones, while the last two bi-syllabic words, 非典 and 沙士, are of similar functional values, and got chosen by different communities by chance.

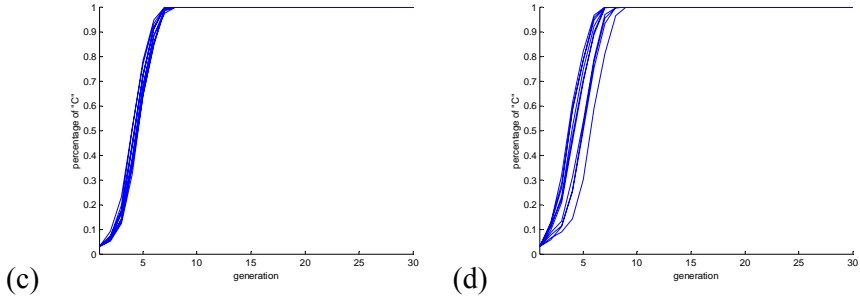

(c)      (d)

Figure 15. Diffusion dynamics in four types of networks in 10 runs. (N=400, <k>=20, β=10, I=10, $α_{adult}$ =0.001). (a) regular network; (b) small-world network (*p=0.01*); (c) random network; (d) scale-free network.

Figure 16 gives the probability of successful diffusion over 100 runs for different functional biases. We can see that for a range of small functional biases (β=3~7), regular and small-world networks have higher probabilities of successful diffusion than scale-free and random networks (Figure 16(a)). In other words, the second two types of networks have a higher threshold of functional bias for successful diffusion. But the second two types of networks take much less time to complete the diffusion than the first two, as shown by average diffusion time over 100 runs in Figure 16(b).

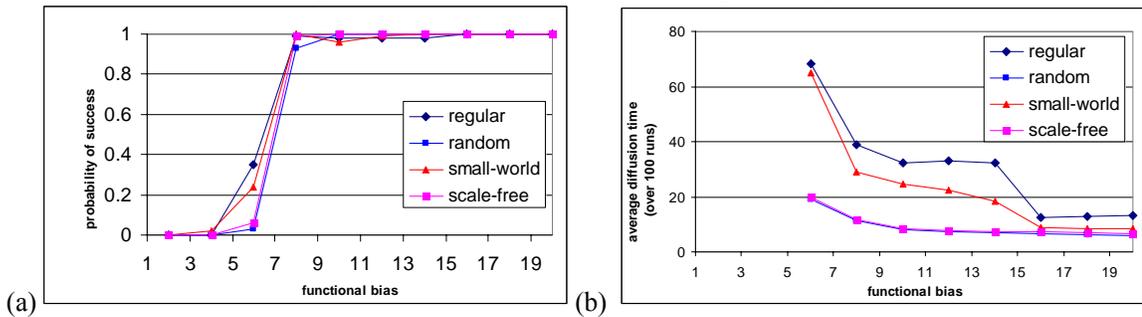

(a)      (b)

Figure 16. (a) Probabilities of successful diffusion under different functional biases; (b) Average diffusion time over 100 runs (N=400, <k>=20, I=10).

When either the functional bias (β) or the number of innovators (I) is high, there are little differences between the four types of networks. However, in most cases of the real world, innovations do not have very high functional biases, sometime even no bias; also they often only occur in a small number of innovators except for massive contact situations. For a range of small functional bias and small number of innovators, as shown in Figure 16, β∈[4,8], the four types of networks show different characteristics: the dynamics in small-world networks is similar to that in regular networks: high success probability, but slow diffusion rate. Scale-free networks are similar to random networks: fast diffusion rate, but lower success probability. Though the range of functional bias to show the differences may still be unrealistically high, and the differences between the two types of networks may not be significant, it is worthwhile to explore the potential differences of these types of networks.

### 3.6 Discussion for the models of language change and future works

We have presented a series of models for the study of language change, simulating change as an innovation diffusion process. By gradually removing some constraints in highly idealized models, such as infinite population size, immortal agents, homogeneous population without any social structure, we add more and more realistic considerations into the model: the agents have a limited life span and progress through five age stages; they have different learning capacities at different age stages; they have connections with only a proportion of the community; they learn from the connected teachers and adopt the form which has more impact in the environment. And the model allows manipulations of several parameters, for example, the degree of connectivity, the learnability of the innovation, the overall structure of networks, etc., to study their effect on the diffusion dynamics. This process of refining the model gives a demonstration of how a computational model for language change can be developed and explored.

With respect to the role of social network on language change, we have examined four typical types of networks and their effect on the diffusion dynamics. When the functional bias is sufficiently high, innovations always spread in a linear manner in *regular* and *small-world* networks, but diffuse quickly in a sharp S-curve in *random* and *scale-free* networks. The success rate of diffusion is higher in regular and small-world networks, but the diffusion rate is much higher in random and scale-free networks. These two types of dynamics lead to questions both for modeling and for empirical studies.

On the one hand, for modeling, there is a question: which types of networks are more appropriate to model the population structure? Under some conditions, the four types of networks give two types of diffusion dynamics. Then, which type of network is more realistic, the small-world or the scale-free network? Or is it necessary to develop a new type of network model which can incorporate both the small-world and scale-free features? The difficulty is that the model needs to accommodate both the presence of a large proportion of local connections for the majority of agents, and the presence of some hubs in the network. This will be an interesting topic for the studies of network modeling.

On the other hand, the study also raises the question for empirical studies on social networks, in particular for the relation of language change and social networks to explain historical data — what would the social network look like at a certain historical period for a community? So far there have been few data of large scale social networks with linguistic behavioral data. This question is of particular interest in the various situations of language contact. More systematic analyses of the social networks of those contact situations may provide useful data for network modeling to build upon.

There are abundant data from historical linguistics on various contact situations. For example, the history of English provides two distinctive situations of language contact: 1) a large number of immigrants with low social impact, as in the settlement of a large number of Scandinavians in the north of England for about 250 years, which resulted in only a handful of lexical borrowings; 2) a small group of immigrants with high prestige, as in the case of French influence on English for about 300 years, which led to the replacement of about one third of the English vocabulary with French words. The model will be useful in carrying out controlled experiments on manipulating various conditions, such as the ratios of population size between two groups in contact, contact intensity and duration, the presence of differentiated social status, etc. It will be interesting to see if the model can predict such two kinds of different outcomes, and compare them with the empirical data.

Though the models presented so far are still simple and highly idealized, they can be used to address some empirical questions in historical linguistics as well as sociolinguistics. For instance, there are two existing hypotheses from empirical sociolinguistic studies regarding the leaders of language change. Some suggest that leaders of changes are centrally located in social networks (Labov 2001), while some remark that marginal members with weak ties within the community are leaders (Milroy 1980/1987). The controversy may be resolved by running simulations of different conditions where innovation arise from the two types of speakers (center or marginal members), and compare the success rates of diffusion in these conditions. Our earlier simulation has shown that when an agent with a large number of connections adopts the innovation, the diffusion starts to pick up speed and spread very quickly into the population (Ke et al. 2004). This may suggest that the central members of the population, who are highly connected, play more important roles in the diffusion process. More systematic experimentation may help to provide a better answer to this controversy.

The consideration of social networks is important not only in the study of language change, but also in modeling language emergence at a macro time scale (Gong et al. 2004). A possible scenario for language emergence is that a number of innovations of more complex structures are created by a small number of innovative persons, and others in the population and in the next generation learn these innovations (and may create their own innovative language expressions in turn as inspired by these learned innovations).

During the process of diffusion of linguistic innovations in small communities of early humans[14], social networks may have played an important role as well.

## 4. General discussions

According to Newman (2003), there are three main directions in network research. First, analyze the statistical properties which characterize the structure and behavior of networked systems, and suggest appropriate ways to measure these properties. Second, create models of networks that can help us to understand the meaning of these properties and explain how they came into being and how they interact with each other. Third, predict what the behavior of networks will be on the basis of certain structural properties, and study the effect of the network structures on the system behavior.

What has been reported in the first part of the paper, networks in the language system, exemplifies research in the first research direction. The analyses of networks have become more and more elaborate, and more complex measures have been proposed, such as centrality, navigation efficiency, information flow, motifs, and so on. The analyses of directed and weighted networks are also receiving more attention as well. With the development of these techniques, it is expected that more complex linguistic networks can be constructed and analyzed, and provide a new paradigm to study the organization and evolution of language, both for language in the individual speakers and language in the speech communities.

The study of the dynamics of language change in simulation models with population in network structures, which is reported in the second part of the paper, contributes as a rich area for exploration in the third research direction. The models allow systematic studies on how different network structures affect the diffusion dynamics and language change in the long run, and provide a computational framework to address some theoretical questions in historical linguistics, for example, if the rate of language change is constant (Nettle 1999b), if change from above and below lead to different types of change (Labov 1994), how linguistic diversity arises and if it is predictable by some parameters such as population size (Nettle 1999c), and so on.

Both the studies on linguistic networks and social networks raise questions for the second research area — new models of network, which can reflect better the features of these real-world networks. The current two typical network models, i.e. small-world and scale-free networks, each can capture some of the features of realistic networks, but it is necessary to explore models which can incorporate features from both models (a recent model reported in Li & Chen (2003) which proposes a local-world view to grow the network seems a promising model). And there are also other features, such as the presence of hierarchical structures and local communities, the multiplex links among individuals, that all need to be taken into account in future models.

All in all, network research will bring a new perspective to linguistics, provide a new methodology to carry out quantitative analyses and suggest new questions and insights; at the same time, studies of networks about language will bring up new challenges to network research in general, and enrich the field with abundance of empirical data and questions. This is a cross-fertilization area worthy exploring.

---

[14] Dunbar (1996) proposed that the size of the population at the time human language emerged would be about 50-100, conjecturing from his studies on many primate societies and hunter-gathering communities.

## Appendix

A brief glossary to some simple measures of network properties[15]:

Measures of the individual nodes:
- degree ($k_i$) : the number of links the node *i* has. Nodes with a large number of links are considered as "hubs" in an undirected network, or "hubs" and "authorities" in a directed network.
- distance ($d_{ij}$) between node *i* to node *j* : the number of links along the shortest path to connect the two nodes. If two nodes are very distant from each other, it needs go through a large number of intermediate nodes and links.
- clustering coefficient ($C_i$) : measures how well a node's neighbors are connected to each other. If a node has a high C, it means the node is within a cluster with dense links.
- centrality ($g_i$) : often measured as the betweenness centrality, which counts how many shortest paths go through the node. The nodes having a high centrality usually exist as connectors between clusters in the network.

Measures of the global statistical properties of the whole network:
- **connectivity**: the average degree of the nodes <*k*>. It is one indicator of the **density** of the network's connection. Usually real-world networks are sparse, which means the average number of links per node is typically smaller than the total number of nodes in the network.
- **degree distribution:** the statistical distribution of the degrees of the nodes. Usually it is plotted as a histogram or extrapolated as a distribution function **P(*k*)**. It gives an idea of the homogeneity and scaling properties of a network. Regular networks are homogeneous as the degrees of all nodes are the same, and its P(*k*) is a delta function shown as a single spike in the histogram. Typical random networks are found to follow a Poisson distribution. Scale-free networks show a power-law degree distribution.
- **characteristic path length** (also called **average path length, L**): the average shortest distance between any two nodes. This gives an idea of the effective size of the network. For regular networks, L increases linearly with respect to the size of the network. On the contrary, in random networks, the distance between any two nodes is small even the network is of huge size; L increases logarithmically, rather than linearly, when the network size increases. Small-world and scale free networks share this feature with random networks.
- **clustering coefficient (C)**: the average clustering coefficient of all nodes. Random networks have a small C, and tend to zero in the limit of large network size. On the contrary, regular networks and small-world networks are highly clustered compared to random networks.
- **assortativeness** (**Γ**): how likely nodes of similar characteristics are connected to each other. A network is said to show assortative mixing if the nodes which have many links tend to be connected to other nodes with many links. The opposite is called disassortative mixing.

---

[15] The actual implementation of the measure for these properties involves many detailed considerations. More rigorous definitions and discussions on various implementations of the above measures can be found in Newman (2003).